\documentclass[11pt,a4paper]{article}
\usepackage[hyperref]{acl2020}
\usepackage{times}
\usepackage{amsmath}
\usepackage{booktabs}
\usepackage{graphics}
\usepackage{latexsym}
\usepackage{nidanfloat}
\usepackage{enumitem}

\usepackage{microtype}

\usepackage{xargs}
\usepackage[colorinlistoftodos,prependcaption,textsize=tiny]{todonotes}

\definecolor{Salmon}{RGB}{250,128,114}
\newcommandx{\aditya}[2][1=]{\todo[linecolor=Salmon,backgroundcolor=Salmon!25,bordercolor=Salmon,#1]{#2}}

\definecolor{DeepSkyBlue}{RGB}{0,191,255}
\newcommandx{\maryam}[2][1=]{\todo[linecolor=DeepSkyBlue,backgroundcolor=DeepSkyBlue!25,bordercolor=DeepSkyBlue,#1]{#2}}

\definecolor{GreenYellow}{RGB}{173,255,47}
\newcommandx{\marshall}[2][1=]{\todo[linecolor=GreenYellow,backgroundcolor=GreenYellow!25,bordercolor=GreenYellow,#1]{#2}}

\aclfinalcopy 

\title{Data Augmentation for Training Dialog Models Robust to \\Speech Recognition Errors}

\author{Longshaokan Wang \; Maryam Fazel-zarandi \; Aditya Tiwari \\
\textbf{Spyros Matsoukas \; Lazaros Polymenakos} \\
Alexa AI, Amazon \\
\texttt{\{longsha, fazelzar, aditiwar, matsouka, polyml\}@amazon.com} \\}

\date{}

\begin{document}

\maketitle
\begin{abstract}
Speech-based virtual assistants, such as Amazon Alexa, Google assistant, and Apple Siri, 
typically convert users' audio signals to text data through automatic speech recognition (ASR) 
and feed the text to downstream dialog models for natural language understanding and response generation. 
The ASR output is error-prone; however, the downstream dialog models are often trained on error-free text data, 
making them sensitive to ASR errors during inference time. 
To bridge the gap and make dialog models more robust to ASR errors, we leverage an ASR error simulator to inject noise into the error-free text data, 
and subsequently train the dialog models with the augmented data. 
Compared to other approaches for handling ASR errors, such as using ASR lattice or end-to-end methods, 
our data augmentation approach does not require any modification to the ASR or downstream dialog models; 
our approach also does not introduce any additional latency during inference time. 
We perform extensive experiments on benchmark data and show that our approach 
improves the performance of downstream dialog models in the presence of ASR errors, 
and it is particularly effective in the low-resource situations where there are constraints on model size or the training data is scarce.
\end{abstract}

\section{Introduction}

Speech-based virtual assistants, such as Amazon Alexa, Google Assistant, and Apple Siri, 
have become increasingly powerful and popular in our everyday lives, 
offering a wide range of functionality including controlling smart home devices, booking movie tickets, and even chit-chatting. 
These speech-based virtual assistants typically contain the following components: 
an automatic speech recognition (ASR) module that converts audio signals from a user to a sequence of words, 
a natural language understanding (NLU) module that extracts semantic meaning from the user utterance, 
a dialog management (DM) module that controls the dialog flow and communicates with external applications if necessary,
a natural language generation (NLG) module that converts the system response to natural language,  
and a text-to-speech (TTS) module that converts the text response to an audio response \cite{Jurafsky2009}. 
The errors made by the ASR module can propagate to the downstream dialog models 
in NLU and DM and degrade their performances \cite{Serdyuk2018, Shivakumar2019a}. 

One straightforward approach to improve the downstream dialog models' robustness to ASR errors is to train them 
with ASR hypotheses with potential ASR errors in addition to the error-free reference texts. 
However, the training data might not always have corresponding ASR hypotheses available, 
for example, when the training data are created in written forms from the beginning. 
Such training data include online reviews, forums, and data collected in a Wizard-of-Oz (WOZ) setting \cite{Rieser2011}.
Additionally, even when there are ASR hypotheses available, transcribing and annotating the ASR hypotheses 
to create the training data is a slow and expensive process due to human involvement, limiting the size of available training data. 

To address these challenges, we propose a simple data augmentation method 
leveraging a confusion-matrix-based ASR error simulator \cite{FazelZarandi2019, Schatzmann2007}. 
Our method can be used on training data with or without existing ASR hypotheses, 
does not require modifying the ASR model or downstream dialog models, 
and consequently does not introduce additional latency during inference time. 
We assess the method's effectiveness on a multi-label classification task on a public dataset from DSTC2 \cite{Henderson2014}. 
We show that our method can improve the dialog models' performances in the presence of ASR errors, 
particularly in the low resource situations where there are model size or latency constraints, or the training data is scarce. 

\section{Related Work}

Existing approaches for handling ASR errors generally fall into four categories: 
1) pre-training word embeddings such that acoustically similar words are close to each other in the embedding space 
\cite{Shivakumar2019a, Shivakumar2019b, Ghannay2016};
2) using multi-task training to jointly correct ASR errors in addition to performing the original NLU tasks
\cite{Schumann2018, Weng2020};
3) using n-best ASR hypotheses, word confusion networks, 
or ASR lattice produced by the ASR system as the input for the downstream dialog models, 
allowing the models to consider all the alternatives instead of only the 1-best ASR hypothesis 
\cite{Weng2020, Ladhak2016, Dilek2006}; and 
4) using an end-to-end approach that combines ASR and NLU systems into one, 
extracting semantics directly from audio signals 
\cite{Serdyuk2018, Haghani2018}. 
These approaches often either require significant modifications to ASR model and/or downstream dialog models, 
or require access to additional information from the ASR model, such as ASR n-best or ASR lattice, during inference time. 
In comparison, data augmentation is much simpler because 
it does not modify the existing model architecture or introduce additional latency during inference time. 
Data augmentation has a long history in image processing \cite{Shorten2019}. 
In language processing, researchers have proposed back-translation \cite{Einolghozati2019} 
and simple operations such as synonym replacement and random swap \cite{Wei2019} to increase the variations of training data. 
These data augmentation approaches aim to improve dialog models' robustness to surface form variations in general, 
whereas our approach focuses on robustness to ASR errors in particular. 
Note that our data augmentation approach can be complimentary to using acoustic embeddings (first category), multi-task training (second category), 
and the other mentioned data augmentation approaches, 
and it is possible to combine them for further performance gains. 

\section{Method}
\label{sec:method}

We propose to use simulated ASR hypotheses to augment the training data of dialog models. 
To this end, we adopt the confusion-matrix-based ASR error simulator 
initially proposed by \citeauthor{Schatzmann2007} (\citeyear{Schatzmann2007}) and improved by \citeauthor{FazelZarandi2019} (\citeyear{FazelZarandi2019}). 
Here we describe the error simulator at a high level, while leaving the details to the mentioned references. 
The main component of the error simulator is an n-gram confusion matrix 
constructed from a corpus of ASR hypotheses and corresponding reference texts: 
Each ASR hypothesis and its reference text are aligned at the word level by minimizing the Levenshtein distance between them, 
then the frequencies of n-gram confusions are added to the confusion matrix 
for $n \in [1, M]$, where $M$ is a pre-specified constant.
During inference time, the error simulator first partitions a reference text into n-grams where $n$ can vary, 
then for each n-gram it samples a replacement from the confusion matrix with 
sampling probabilities proportional to the frequencies of confusions. 
Note that the sampled ``confusion'' can be the original n-gram itself, 
which means correct recognition for this n-gram in the simulated hypothesis. 

We refer to the corpus used to construct the n-gram confusion matrix as the \textsl{ASR corpus} 
to distinguish it from the training data for the dialog models that we want to apply the error simulator to. 
By design, if the reference texts that the error simulator is applied to have the same distribution 
as the reference texts in the \textsl{ASR corpus}, 
then the simulated ASR hypotheses will have the same error distribution  
as the ASR hypotheses in the \textsl{ASR corpus} \cite{Schatzmann2007}, 
where the error distribution includes word-error-rate (WER) and proportions of insertion, deletion, and substitution errors.
However, in practice it can be useful to simulate ASR hypotheses with 
a pre-specified WER different from that of the \textsl{ASR corpus}. 
Adjusting the WER is non-trivial, because each word's individual WER is often different 
from the overall WER of the \textsl{ASR corpus}; i.e., some words are more easily confusable than others. 
We introduce a heuristic to adjust each word's individual WER during inference time of the error simulator so that the 
overall WER in the simulated ASR hypotheses is close to the pre-specified target overall WER based 
on the following formula (see Appendix \ref{appendix:wer}):
$$
\resizebox{\columnwidth}{!}{$\displaystyle{\frac{1 - \text{target individual WER}}{1 - \text{original individual WER}} = \frac{1 - \text{target overall WER}}{1 - \text{original overall WER}}.}$}
$$
This heuristic has the following desired properties: 
1) If $w_1$ has a higher original individual WER than that of $w_2$ before the adjustment, 
then $w_1$ will have a higher target individual WER than that of $w_2$ from this adjustment, 
for arbitrary words $w_1$ and $w_2$ under certain simplifying conditions (Appendix \ref{appendix:wer}); 
i.e., we mostly preserve the property that some words are more easily confusable than others. 
2) In the trivial case where all words have the same individual WER as the overall WER, 
this heuristic is equivalent to setting all individual WER to the target overall WER.

We apply data augmentation to the training data of dialog models in two different cases: 
\textbf{S1}) The training data only have reference texts and no corresponding ASR hypotheses. 
In this case, we construct the confusion matrix used by the ASR error simulator with an \textsl{ASR corpus} ideally close to 
the training data of dialog models in terms of vocabulary overlap, 
or a large generic \textsl{ASR corpus} such as Fisher English Training Speech Corpora \cite{Cieri2004, Shivakumar2019b}.
We simulate multiple ASR hypotheses for each reference sentence with different WER, 
and combine all the simulated ASR hypotheses with the reference text as the augmented training data -- 
the motivation behind this is to create more variations, 
make dialog models robust to different levels of WER, 
and avoid degradation on error-free data. 
\textbf{S2}) The training data have both reference texts and corresponding ASR hypotheses. 
In this case, we can directly use the training data as the \textsl{ASR corpus} to construct the confusion matrix, 
then simulate ASR hypotheses with different WER and combine them with the 
original ASR hypotheses and reference texts as the augmented training data. 
Note that during inference time of the ASR error simulator, it partitions a sentence and samples n-gram replacements probabilistically, 
so even though we use the training data of dialog models as the \textsl{ASR corpus}, 
the error simulator can still create new variations in the simulated ASR hypotheses.

\section{Experiments}

We experiment our proposed data augmentation method on the dialog act classification task.

\subsection{Data}
\label{subsec:data}

We use the public dataset from DSTC2 \cite{Henderson2014}, which has reference texts, ASR hypotheses, and dialog act annotations.
We choose DSTC2 because the other commonly used NLU datasets often don't have ASR hypotheses available. 
This dataset consists of human-computer dialogs in a restaurant domain collected using Amazon Mechanical Turk. 
We follow the same data preprocessing steps as in \citeauthor{Weng2020}'s (\citeyear{Weng2020}) work. 
After preprocessing, the dataset has 10,876/3,553/9,153 training/validation/test samples and $25$ unique dialog act labels. 
Each user utterance may have multiple dialog act labels, thus we treat this problem as a multi-label classification problem. 
More specifically, in the training set, 7,516 utterances have $1$ dialog act label each, 
3,254 utterances have $2$ dialog act labels each, and $106$ utterances have $3$ dialog act labels each. 
The ASR hypotheses for the user utterances have a WER of $27.89\%$, 
where the errors consist of $58.96\%$ substitutions, $15.66\%$ insertions, and $25.38\%$ deletions.
Additionally, in $45\%$ of the test cases, the ASR hypothesis has perfect recognition.

\subsection{Setup}
\label{subsec:setup}

We measure the effectiveness of data augmentation in both cases mentioned in Section \ref{sec:method}. 
In the first use case, we assume that the training and validation sets have no ASR hypotheses, 
and we need to construct the n-gram confusion matrix with a separate \textsl{ASR corpus}. 
The \textsl{ASR corpus} used for constructing the confusion matrix for the error simulator is a separate dataset of 
10,000 transcribed utterances from different domains such as movie recommendation, ticket booking, and restaurant booking.
As a measure of similarity, this \textsl{ASR corpus} contains $43.3\%$ of unique words, $12.6\%$ of unique bigrams, and $4.4\%$ of unique trigrams from DSTC2.
In addition to experimenting with data augmentation with the confusion-matrix-based ASR error simulator, 
we consider data augmentation with a much simpler error simulator which we call \textsl{the uniform error simulator}, 
to see whether a simpler error simulator would suffice. 
The uniform error simulator samples word replacements from the training data vocabulary uniformly with a pre-specified WER. 
The training and validation data in each setting are as follows:
\begin{itemize}[leftmargin=0.4in]
\item[S1-1] Reference utterances only (baseline);
\item[S1-2] Reference utterances + simulated ASR hypotheses by the uniform error simulator with $27.9\%$ WER;
\item[S1-3] Reference utterances + simulated ASR hypotheses by the confusion-matrix-based error simulator with $27.9\%$ WER\footnote{
We use $27.9\%$ WER to match the WER of DSTC2. 
Note that in this case even though we assume that we don't have ASR hypotheses in the training data, 
we may still know the overall WER of the ASR system. If the overall WER is unknown, we can always use a combination of different WER similar to S1-4.};
\item[S1-4] Reference utterances + three sets of simulated ASR hypotheses by the confusion-matrix-based error simulator with $27.9\%$, $20\%$, $15\%$ WER, respectively.
\end{itemize}
Note that the ratios between reference utterances and simulated hypotheses are 1:1 for S1-2 and S1-3 and 1:3 for S1-4.

In the second use case, the training and validation sets have both reference utterances and ASR hypotheses. 
We use the training set itself as the \textsl{ASR corpus} to construct the confusion matrix.
Compared to the first case, we include the original ASR hypotheses in the training and validation sets for each setting, 
but we do not include the simulated ASR hypotheses in the validation set to keep this set as close to the test set as possible.
We refer to the settings for the second case as S2-1, S2-2, S2-3 and S2-4.

\begin{table*}[t]
  \caption{Accuracy (\%) and F1-score results on ASR hypotheses and reference texts 
  from the DSTC2 test set for different settings described in Section \ref{subsec:setup}.
  ``Uniform'' refers to the uniform error simulator, and ``Conf. Mat.'' refers to the confusion-matrix-based error simulator.}
  \label{tab:ic-best}
  \centering
    \resizebox{2.1\columnwidth}{!}{
  \begin{tabular}{cllcccc}
    \toprule
    &
    \multicolumn{1}{c}{\textbf{Training Setup}} & 
    \multicolumn{1}{c}{\textbf{Validation Setup}} & 
     \multicolumn{2}{c}{\textbf{Hypothesis}} &
    \multicolumn{2}{c}{\textbf{Reference}} \\
    & & &
    \multicolumn{1}{c}{\textbf{Accuracy}} & 
    \multicolumn{1}{c}{\textbf{F1-Score}} &
    \multicolumn{1}{c}{\textbf{Accuracy}} & 
    \multicolumn{1}{c}{\textbf{F1-Score}} \\
    \midrule
    S1-1 & ref & ref & $80.76$ & $	0.8935$& $96.98$ & $0.9847$   \\
    S1-2 & ref + sim hyp (Uniform, $27.9\%$ WER) & ref + sim hyp (Uniform, $27.9\%$ WER) & $80.80$ & $0.8938$ & $96.36$ & $0.9814$   \\
    S1-3 & ref + sim hyp (Conf. Mat., $27.9\%$ WER)  & ref + sim hyp (Conf. Mat., $27.9\%$ WER) & $80.98$ & $0.8950$ & $96.95$ & $0.9845$   \\
    S1-4 & ref + sim hyp (Conf. Mat., mixed WER) & ref + sim hyp (Conf. Mat., mixed WER) & $81.05$ & $0.8953$ &$97.19$ & $0.9857$   \\
    \midrule
    S2-1 & ref + hyp & ref + hyp & $83.03$ & $0.9073$ &  $96.88$ & $0.9841$   \\
    S2-2 & ref + hyp + sim hyp (Uniform, $27.9\%$ WER) & ref + hyp & $82.73$ & $0.9054$ &  $96.49$ & $0.9821$   \\
    S2-3 & ref + hyp + sim hyp (Conf. Mat., $27.9\%$ WER) & ref + hyp & $83.16$ & $0.9080$ & $96.59$ & $0.9827$   \\
    S2-4 & ref + hyp + sim hyp (Conf. Mat., mixed WER) & ref + hyp & $83.19$ & $0.9082$ & $96.80$ & $0.9837$  \\
    \bottomrule
  \end{tabular}
    }
\end{table*}

We use the FLAIR package \cite{Akbik2018} to build the multi-label classifier, 
tune the hyperparameters on the validation set with hyperopt \cite{hyperopt} under the baseline setting S1-1, 
and keep the same model architecture across all settings. 
The final model architecture selected by hyperopt is BERT embeddings \cite{Devlin2019} + 
embedding re-projection + $1$ layer bidirectional LSTM with $256$ hidden size. 
In each setting, we use the validation set for learning rate decay and early stopping, 
and test the trained model on the ASR hypotheses and reference utterances in the test set separately. 
We measure the model performance with micro-averaged accuracy and F1-score. 
Because each sample only has one or a few class labels out of the $25$ possible labels, 
we follow FLAIR's convention in calculating the accuracy for multi-label classification by 
excluding the number of true negatives from both the numerator and denominator:
$$
\resizebox{\columnwidth}{!}{$\displaystyle{\text{Accuracy} = \frac{(\text{\# true pos.})}{(\text{\# true pos.}) + (\text{\# false pos.}) + (\text{\# false neg.})}.}$}
$$

\subsection{Results}
\label{subsec:results}

The results are shown in Table \ref{tab:ic-best}. We see that in both cases, 
the proposed data augmentation settings (S1-4 and S2-4) yield the best performance when tested on ASR hypotheses. 
However, the improvements over the baselines (S1-1 and S2-1) are very marginal ($0.29\%$ absolute improvement for S1 and $0.16\%$ for S2), 
which could be due to the baseline performances already being close to optimal,
or data augmentation being less effective for models with powerful pre-trained embeddings like BERT, 
which \citeauthor{Shleifer2019} (\citeyear{Shleifer2019}) also observed in his work. 
For S1-4, we also experiment with using the DSTC2 ASR hypotheses as the \textsl{ASR corpus} as an oracle/upper bound setting for S1,
where we assume that the training data have no ASR hypotheses but the \textsl{ASR corpus} is perfectly in-domain, which yields
an accuracy of $82.90\%$ testing on hypotheses and $96.87\%$ on references.

Examining the errors made by S1-1, S1-4, S2-1, and S2-4 settings on the ASR hypotheses, 
we see that more than $90\%$ of the errors are common among the four settings.
Furthermore, in about $54\%$ of the cases there is no overlap between the words in the references and the corresponding ASR hypotheses.
For example, the user utterance ``Danish'' is recognized as ``the address'', and ``seafood'' is recognized as ``is serve''.
This indicates that these ASR hypotheses are distorted to the extent that it would be very unlikely to correctly predict the label, 
even for a human annotator.
In an additional $32\%$ of the common errors, the ASR hypotheses have changed the semantics of the reference texts.
For example, ``Spanish food'' is recognized as ``which food'', ``no Thai'' is recognized as ``no hi'', and ``no Italian food'' is recognized as ``would like Italian food''. 
In these cases, it would also be almost impossible to predict the ground-truth labels. 

\begin{table}[t]
  \caption{Accuracy (\%) and F1-score results for reduced model architectures. 
  Except for model architecture, the settings are exactly the same as those of Table \ref{tab:ic-best}. 
  ``without BERT'' refers to the model architecture from Table \ref{tab:ic-best} with BERT replaced with GloVe embeddings; 
  ``Simple NN'' refers to the GloVe embeddings + $1$ layer unidirectional LSTM with $128$ hidden size architecture.}
  \label{tab:ic-simple}
  \centering
  \resizebox{\columnwidth}{!}{
  \begin{tabular}{clcccc}
    \toprule
    \multicolumn{1}{c}{\textbf{Training}} & 
    \multicolumn{1}{c}{\textbf{Model}} &
    \multicolumn{2}{c}{\textbf{Hypothesis}} &
    \multicolumn{2}{c}{\textbf{Reference}}  \\
    \multicolumn{1}{c}{\textbf{Setup}} & 
    \multicolumn{1}{c}{\textbf{Architecture}} &
    \multicolumn{1}{c}{\textbf{Acc.}} & 
    \multicolumn{1}{c}{\textbf{F1}} &
    \multicolumn{1}{c}{\textbf{Acc.}} & 
    \multicolumn{1}{c}{\textbf{F1}} \\
    \midrule
     S1-1 & without BERT & $80.34$ & $0.8910$ & $94.09$ & $0.9695$   \\
     S1-2 & without BERT & $81.12$ & $0.8958$ & $94.55$ & $0.9720$  \\
     S1-3 & without BERT & $80.81$ & $0.8939$ & $94.33$ & $0.9709$  \\
     S1-4 & without BERT & $81.19$ & $0.8962$ & $95.36$ & $0.9762$ \\
     \midrule
     S2-1 & without BERT & $80.40$ & $0.8913$ & $92.18$ & $0.9593$   \\
     S2-2 & without BERT & $81.47$ & $0.8979$ & $93.88$ & $0.9685$ \\
     S2-3 & without BERT & $81.32$ & $0.8969$ & $93.24$ & $0.9650$  \\
     S2-4 & without BERT & $82.63$ & $0.9049$ & $95.14$ & $0.9751$  \\
    \midrule
     S1-1 & Simple NN & $76.92$ & $0.8695$  & $87.74$ & $0.9347$  \\
     S1-2 & Simple NN  & $77.01$ & $0.8701$ & $87.11	$ & $0.9311$\\
     S1-3 & Simple NN & $79.66$ & $0.8868$ & $91.82$ & $0.9573$  \\
     S1-4 & Simple NN & $80.81$ & $0.8939$ & $94.28$ & $0.9706$ \\
     \midrule
     S2-1 & Simple NN & $79.35$ & $0.8848$ & $89.65$ & $0.9454$   \\
     S2-2 & Simple NN & $79.65$ & $0.8867$ & $90.63$ & $0.9508$ \\
     S2-3 & Simple NN & $81.76$ & $0.8996$ & $93.87$ & $0.9684$  \\
     S2-4 & Simple NN & $81.69$ & $0.8992$ &  $94.16$ & $0.9699$ \\     
    \bottomrule
  \end{tabular}
  }
\end{table}

\begin{table*}[b]
  \caption{Accuracy (\%) results for training on different sub-samples of the data with different model architecture  on the hypothesis test set. 
  We compare the baselines (S1-1 and S2-1) with the proposed data augmentation settings (S1-4 and S2-4) 
  and show the absolute improvements in accuracy.}
  \label{tab:ic-splits-hyp}
  \centering
    \resizebox{1.35\columnwidth}{!}{
  \begin{tabular}{lccccccc}
    \toprule
    \multicolumn{1}{c}{\textbf{Model}} & 
    \multicolumn{1}{c}{\textbf{Subsample}} & 
	\multicolumn{3}{c}{\textbf{S1 Setting}} & 
	\multicolumn{3}{c}{\textbf{S2 Setting}} \\
	 \multicolumn{1}{c}{\textbf{Architecture}} &
	\multicolumn{1}{c}{\textbf{Proportion}} &
	\multicolumn{1}{c}{\textbf{S1-1}} &
	\multicolumn{1}{c}{\textbf{S1-4}} &
	\multicolumn{1}{c}{\textbf{Gain}} &
	\multicolumn{1}{c}{\textbf{S2-1}} &
	\multicolumn{1}{c}{\textbf{S2-4}} &
	\multicolumn{1}{c}{\textbf{Gain}}  \\
    \midrule
    With BERT& $1\%$ &$43.67$ & $64.94$ & $+21.27$ & $61.81$ & $62.32$ & $+0.51$ \\
	&$5\%$ &$71.79$ & $77.65$ & $+5.86$ & $74.51$ & $78.62$ & $+4.11$ \\
	&$10\%$ &$78.29$ & $79.41$ & $+1.12$ & $79.90$ & $80.23$ & $+0.33$ \\
	&$25\%$ &$79.37$ & $80.15$ & $+0.78$ & $80.88$ & $81.56$ & $+0.68$ \\
	&$50\%$ &$80.54$ & $80.8$ & $+0.26$ & $82.00$ & $82.40$ & $+0.40$ \\
	&$75\%$ & $81.05$ & $81.27$ & $+0.22$ & $82.58$ & $83.16$ & $+0.58$\\
	\midrule
	Without BERT & $1\%$ &$5.72$ & $4.48$ & $-1.24$ & $6.17$ & $0.81$ & $-5.36$\\
	&$5\%$ &$0.88$ & $73.01$ & $+72.13$ & $0.02$ & $70.33$ & $+70.31$ \\
	&$10\%$ & $0.12$ & $76.16$ & $+76.04$ & $69.91$ & $76.65$ & $+6.74$\\
	&$25\%$ &$70.75$ & $78.47$ & $+7.72$ & $76.71$ & $80.55$ & $+3.84$ \\
	&$50\%$ &  $78.73$ & $80.59$ & $+1.86$ & $80.15$ & $82.10$ & $+1.95$\\
	&$75\%$ & $78.04$ & $81.38$ & $+3.34$ & $80.92$ & $82.11$ & $+1.19$\\
	\midrule
	Simple NN & $1\%$ &$15.90$ &$16.68$ &$+0.78$ &$16.55$ &$15.59$ &$-0.96$ \\
	&$5\%$ &$11.11$ &$41.10$ &$+29.99$ &$8.28$ &$41.47$ &$+33.19$ \\
	&$10\%$ &$8.55$ &$74.10$ &$+65.55$ &$0.07$ &$72.61$ &$+72.54$ \\
	&$25\%$ &$12.11$ &$77.37$ &$+65.26$ &$75.84$ &$77.70$ &$+1.90$ \\
	&$50\%$ &$75.72$ &$80.08$ &$+4.36$ &$76.99$ &$79.54$ &$+2.55$ \\
	&$75\%$ & $76.69$ &$80.54$ &$+3.85$ &$78.32$ &$81.53$ &$+3.21$ \\	
    \bottomrule
  \end{tabular}
}
\end{table*}

\subsection{Follow-up Experiments}
\label{subsec:followup}

To investigate whether data augmentation is more effective for models without large pre-trained embeddings like BERT,
we run one set of follow-up experiments where we replace BERT with GloVe embeddings \cite{Pennington2014} 
from the previous model architecture (a $99.79\%$ reduction in number of parameters), 
and another set where we further reduce the model architecture to a much simpler one: 
GloVe embeddings + $1$ layer unidirectional LSTM with $128$ hidden size (a further $83.05\%$ reduction in number of parameters). 
The results are shown in Table \ref{tab:ic-simple}. We see that for the reduced model architectures, 
the improvements from data augmentation are much larger (e.g., $3.89\%$ absolute improvement in S1 and $2.34\%$ in S2 for Simple NN). 
The promising results on simpler model architectures have practical implications, because in certain use cases we may not be able to 
use large model architectures due to constraints on model size, computing power, or latency. 

Lastly, hypothesizing that data augmentation could be particularly effective for limited training data 
even with the best model architecture found by hyperopt, 
we randomly subsample $1\%$, $5\%$, $10\%$, $25\%$, $50\%$, and $75\%$ of the training and validation sets 
(without changing the test sets) and apply the proposed data augmentation with different model architectures.
The comparisons between the baselines (S1-1 and S2-1) and proposed settings (S1-4 and S2-4) on the reduced data 
are shown in Tables \ref{tab:ic-splits-hyp} and \ref{tab:ic-splits-ref}.
As expected, the improvements from data augmentation are generally larger for smaller datasets and simpler model architectures, 
except for most cases on the smallest subset ($1\%$ of training data).

\begin{table*}[t]
  \caption{Accuracy (\%) results for training on different sub-samples of the data with different model architecture on the reference test set. 
  Except for testing on reference utterances instead of ASR hypotheses, the settings are exactly the same as in Table \ref{tab:ic-splits-hyp}.}
  \label{tab:ic-splits-ref}
  \centering
  \resizebox{1.35\columnwidth}{!}{
  \begin{tabular}{lccccccc}
    \toprule
     \multicolumn{1}{c}{\textbf{Model}} & 
    \multicolumn{1}{c}{\textbf{Subsample}} & 
	\multicolumn{3}{c}{\textbf{S1 Setting}} & 
	\multicolumn{3}{c}{\textbf{S2 Setting}} \\
	\multicolumn{1}{c}{\textbf{Architecture}} &
	\multicolumn{1}{c}{\textbf{Proportion}} &
	\multicolumn{1}{c}{\textbf{S1-1}} &
	\multicolumn{1}{c}{\textbf{S1-4}} &
	\multicolumn{1}{c}{\textbf{Gain}} &
	\multicolumn{1}{c}{\textbf{S2-1}} &
	\multicolumn{1}{c}{\textbf{S2-4}} &
	\multicolumn{1}{c}{\textbf{Gain}}  \\
    \midrule
    With BERT& $1\%$ & $53.57$ & $73.75$ & $+20.18$ & $70.30$ & $71.90$ & $+1.60$ \\
    &$5\%$ & $83.58$ & $90.12$ & $+6.54$ & $85.11$ & $90.47$ & $+5.36$ \\
    &$10\%$ & $91.62$ & $92.80$ & $+1.18$ & $91.93$ & $92.99$ & $+1.06$ \\
    &$25\%$ & $93.34$ & $95.49$ & $+2.15$ & $93.84$ & $94.99$ & $+1.15$ \\
    &$50\%$ & $96.01$ & $96.68$ & $+0.67$ & $95.82$ & $95.96$ & $+0.14$ \\
    &$75\%$ & $96.54$ & $96.90$ & $+0.36$ & $96.12$ & $96.49$ & $+0.37$ \\
    	\midrule
	Without BERT & $1\%$ &$5.66$ & $4.38$ & $-1.28$ & $4.35$ & $0.81$ & $-3.54$\\
	&$5\%$ & $0.75$ & $83.56$ & $+82.81$ & $0.04$ & $79.14$ & $+79.10$\\
	&$10\%$ & $0.21$ & $87.03$ & $+86.82$ & $79.08$ & $86.67$ & $+7.59$\\
	&$25\%$ & $81.9$ & $90.25$ & $+8.35$ & $86.85$ & $91.37$ & $+4.52$\\
	&$50\%$ & $90.45$ & $93.77$ & $+3.32$ & $91.27$ & $94.52$ & $+3.25$ \\
	&$75\%$ & $89.7$ & $94.97$ & $+5.27$ & $92.87$ & $94.66$ & $+1.79$ \\
    \midrule
    Simple NN & $1\%$ & $17.44$ & $18.61$ & $+1.17$ & $18.37$ & $16.96$ & $-1.41$ \\
	&$5\%$ & $12.32$ & $45.19$ & $+32.87$ & $10.01$ & $45.21$ & $+35.20$ \\
	&$10\%$ & $8.82$ & $84.83$ & $+76.01$ & $0.07$ & $82.91$ & $+82.84$ \\
	&$25\%$ & $17.55$ & $87.76$ & $+70.21$ & $85.05$ & $87.4$ & $+2.35$ \\
	&$50\%$ & $86.17$ & $92.66$ & $+6.49$ & $86.17$ & $90.17$ & $+4.00$ \\
	&$75\%$ & $88.46$ & $93.88$ & $+5.42$ & $88.16$ & $94.00$ & $+5.84$ \\
    \bottomrule
  \end{tabular}
  }
\end{table*}

\section{Conclusion}

In this paper, we proposed a method for data augmentation in order to make downstream dialog models more robust to ASR errors. 
We leveraged a confusion-matrix-based ASR error simulator to inject noise into the error-free text data, 
and subsequently trained dialog act classification models with the augmented data. 
Compared to other approaches of handling ASR errors, 
our data augmentation approach does not require any modification to the ASR models or downstream dialog models, thus
our approach also does not introduce any additional latency during inference time of the dialog models. 
We performed extensive experiments on benchmark data and showed that our approach 
improves the performance of downstream dialog models in the presence of ASR errors, 
and it is particularly effective in the low-resource situations where the model size needs to be small or the training data is scarce.

For future work, we plan to investigate the effect of our proposed method on additional tasks such as dialog state tracking and response generation.
Additionally, we believe that our data augmentation approach is complimentary to using acoustic embeddings, multi-task training, 
and other mentioned data augmentation approaches, 
and we plan to combine them for further performance gains. 

\section*{Acknowledgments}
We thank Angeliki Metallinou, Sajjad Beygi, Josep Valls Vargas, and the ACL workshop reviewers for the constructive feedback.

\nocite{*}
\bibliography{acl2020}
\bibliographystyle{acl_natbib}

\appendix

\section{Adjusting Word-Error-Rate when Applying the ASR Error Simulator}
\label{appendix:wer}

Please refer to the work by Schatzmann et al. \cite{Schatzmann2007} and Fazel-Zarandi et al. \cite{FazelZarandi2019} 
for details on how the ASR error simulator is applied during inference time, 
which include how out-of-vocabulary words are handled. 
Here we focus on the newly introduced heuristic of adjusting word-error-rate (WER).

Note that a single word can be contained in multiple n-grams in the confusion matrix, 
and when an n-gram is confused with an m-gram for arbitrary $n$, and $m$, 
the actual number of word errors introduced could be greater than, equal to, or less than $n$. 
Thus, making a precise adjustment of a single word's individual WER can be difficult.
Instead, for simplicity, we treat each n-gram in the confusion matrix as a single word for all $n$ when applying this heuristic. 
Under the simplifying condition, the individual WER of a word $w$ can be computed with 
\[
1 - \frac{\text{frequency of correct recognition for } w}{\text{sum of frequencies of confusions for } w},
\] 
where the correct recognition is the same as ``confusing'' with the original word itself. 

Now, assume that we sample from the confusion matrix without any adjustment on the test set when applying the error simulator, 
and the resulting overall WER is $R_1$. 
The target overall WER we want to have is $R_2$. 
We adjust each word's individual WER by changing its frequency of correction recognition based on the following formula:
\[
\resizebox{\columnwidth}{!}{$\displaystyle{\frac{1 - \text{target individual WER}}{1 - \text{original individual WER}} = \frac{1 - \text{target overall WER}}{1 - \text{original overall WER}}.}$}
\]
Then we can derive how many additional correct recognitions, denoted as $X$, to add for a word $w$ as follows:
\begin{align*}
&1 - \text{target individual WER} \\
&= (1 - \text{original individual WER}) \cdot \frac{1 - R_2}{1 - R_1}.
\end{align*}
In practice, the right-hand-side of the above equation can be greater than 1, so we add an upper bound constant $U < 1$:
\begin{align*}
&\resizebox{0.5\columnwidth}{!}{$\displaystyle{1 - \text{target individual WER}}$} \\
&\resizebox{\columnwidth}{!}{$\displaystyle{= \min \left[(1 - \text{original individual WER}) \cdot \frac{1 - R_2}{1 - R_1}, U \right].}$}
\end{align*}
Expanding the original and target individual WER:
\begin{align*}
&\resizebox{0.72\columnwidth}{!}{$\displaystyle{\frac{(\text{freq. of correct recognition for }w) + X}{(\text{sum of freq. of confusions for }w) + X}}$} \\
&\resizebox{\columnwidth}{!}{$\displaystyle{= \min \left[\frac{\text{freq. of correct recognition for } w}{\text{sum of freq. of confusions for } w} \cdot \frac{1 - R_2}{1 - R_1}, U \right].}$}
\end{align*}
$R_1$ and $R_2$ are known, and the frequencies are stored in the confusion matrix, hence we can solve for $X$. Denoting the right-hand-side of the above equation as H, we have:
\begin{align*}
X = [&\resizebox{0.69\columnwidth}{!}{$\displaystyle{H \cdot (\text{sum of freq. of confusions for }w)}$} \\
&\resizebox{0.85\columnwidth}{!}{$\displaystyle{- (\text{freq. of correct recognition for }w)] / (1 - H)}$}.
\end{align*}
Note that $X$ can be negative, which corresponds to having a lower target WER. 

During the inference time of the error simulator, for each word $w$ we sample replacement for, 
we adjust its individual WER by computing $X$ using the last equation and adding it to the frequency of correct recognition for $w$ before sampling.

\end{document}